\documentclass[conference]{IEEEtran}
\usepackage[left=0.625in,right=0.625in,top=0.75in,bottom=1in]{geometry}

\usepackage[numbers]{natbib}
\usepackage{hyperref}
\hypersetup{
    colorlinks=true,
    linkcolor=blue,
    citecolor=blue,
    urlcolor=blue,
}
\usepackage{caption}
\usepackage{subcaption}
\usepackage{multirow}
\usepackage{tabularray}
\usepackage{algorithmic}
\usepackage{graphicx}
\usepackage{textcomp}
\usepackage{fancyhdr}
\usepackage[table]{xcolor} 
\usepackage{colortbl}
\usepackage{makecell}
\usepackage{tabularx}
\usepackage{float}
\usepackage{caption}
\definecolor{PineGreenDarkest}{HTML}{4C9E8F}
\definecolor{PineGreenDark}{HTML}{70B3A1}
\definecolor{PineGreenMedium}{HTML}{94C8B4}
\definecolor{PineGreenLight}{HTML}{B8DDC6}
\definecolor{PineGreenLighter}{HTML}{D1EED9}

\begin{document}
%

\title{Linguistic Intelligence in Large Language Models for Telecommunications}



\author{
    \IEEEauthorblockN{Tasnim Ahmed\IEEEauthorrefmark{3},
    Nicola Piovesan\IEEEauthorrefmark{4}, 
    Antonio De Domenico\IEEEauthorrefmark{4}, 
    Salimur Choudhury\IEEEauthorrefmark{3}}
    
    \IEEEauthorblockA{\IEEEauthorrefmark{3}School of Computing,
    Queen's University, Ontario, Canada}
    
    \IEEEauthorblockA{\IEEEauthorrefmark{4}Paris Research Center,
    Huawei Technologies, Boulogne-Billancourt, France}

    \IEEEauthorblockA{\IEEEauthorrefmark{3}\{tasnim.ahmed, s.choudhury\}@queensu.ca, \IEEEauthorrefmark{4}\{nicola.piovesan, antonio.de.domenico\}@huawei.com}
}


%


\maketitle

\begin{abstract}

Large Language Models (LLMs) have emerged as a significant advancement in the field of Natural Language Processing (NLP), demonstrating remarkable capabilities in language generation and other language-centric tasks. Despite their evaluation across a multitude of analytical and reasoning tasks in various scientific domains, a comprehensive exploration of their knowledge and understanding within the realm of natural language tasks in the telecommunications domain is still needed. This study, therefore, seeks to evaluate the knowledge and understanding capabilities of LLMs within this domain. To achieve this, we conduct an exhaustive zero-shot evaluation of four prominent LLMs—Llama-2, Falcon, Mistral, and Zephyr. These models require fewer resources than ChatGPT, making them suitable for resource-constrained environments. Their performance is compared with state-of-the-art, fine-tuned models. To the best of our knowledge, this is the first work to extensively evaluate and compare the understanding of LLMs across multiple language-centric tasks in this domain. Our evaluation reveals that zero-shot LLMs can achieve performance levels comparable to the current state-of-the-art fine-tuned models. This indicates that pretraining on extensive text corpora equips LLMs with a degree of specialization, even within the telecommunications domain. We also observe that no single LLM consistently outperforms others, and the performance of different LLMs can fluctuate. Although their performance lags behind fine-tuned models, our findings underscore the potential of LLMs as a valuable resource for understanding various aspects of this field that lack large annotated data.
\end{abstract}

%
\IEEEpeerreviewmaketitle

\section{Introduction}
The emergence of Generative Artificial Intelligence (GenAI) technology is viewed as one of the most significant breakthroughs in the era of digital and computational intelligence, revolutionizing how machines understand and generate human-like content. In recent years, the introduction of transformer-based
Large Language Models (LLMs) have led to significant advancements in the field of Natural Language Processing (NLP) across various domains, including telecommunications \cite{10384630, huang2023large}. LLMs can condense information characteristics and transform vast knowledge into tokens, which can assist or even substitute human beings in conceptual comprehension, logical thinking, and decision-making. Language models
have demonstrated state-of-the-art results on various NLP tasks and exhibited significant algorithmic, reasoning, and analytical skills with minimal or no domain-specific adjustments \cite{romera2023mathematical}. This intuitively allows for the effective execution of network tasks via natural language interaction. Consequently, enhancing or optimizing the performance of networking-specific LLMs emerges as an important challenge.

Huang et al. \cite{huang2023large} highlighted the potential uses of LLMs in the networking field. LLMs have the potential to transform network design by analyzing large datasets, aiding in tasks such as equipment selection and network planning. They can also contribute significantly to network diagnosis by using network status data to create fault reports and provide processing suggestions. Moreover, for network configuration, LLMs could provide a unified natural language interface, simplifying the process and assisting in the management of diverse network devices. Finally, LLMs can interface with various security tools and systems, assisting in security evaluation and intrusion detection. Addressing these specialized issues with the help of LLMs necessitates the application of NLP tasks e.g., classification, summarization, question-answering, named entity recognition, relation extraction, etc. However, one major problem is the lack of high-quality datasets in this domain to train these language models. To address this problem, Bariah et al. \cite{bariah2023understanding} proposed a framework for adapting pre-trained generative models, specifically BERT, DistilBERT, RoBERTa, and GPT-2, to the telecom domain. The authors also demonstrated the efficiency of these fine-tuned models in classifying 3rd Generation Partnership Project (3GPP) technical documents into relevant telecom categories and working groups. Another contribution to this domain is the proposal of SPEC5G dataset by Karim et al. \cite{karim2023spec5g}. The authors demonstrated the application of this dataset in tasks like security-related text classification and summarization, aimed at enhancing the understanding and analysis of complex 5G network protocols. In a practical contribution to this research area, Maatouk et al. \cite{maatouk2023teleqna} introduced TeleQnA, a benchmark dataset designed to evaluate the knowledge of LLMs in telecommunications.
In addition, Miao et al. \cite{miao2023empirical} developed the NetEval question-answering dataset, with a focus on network configurations, logs, and events. While supervised fine-tuned language models or zero-shot LLMs have shown impressive results in one or more tasks within the telecommunications field, a thorough evaluation of the capabilities and limitations of LLMs across this domain remains unexplored. 

To this end, we investigate the effectiveness of LLMs in simulating typical NLP tasks in telecommunications research, such as text classification, summarization, and question-answering. In essence, the objective of this study is to enhance our comprehension of the capabilities and limitations of LLMs in the field of telecommunications, which can pave the way for developing new applications in this domain leveraging LLMs. The major contributions of this study are—
\begin{itemize}
    \item A comprehensive zero-shot evaluation of various LLMs within the telecommunications domain, revealing their capabilities and limitations across multiple tasks.
    \item We also carry out a thorough error analysis and provide our observations on the specific areas that need improvement for each of these tasks.
\end{itemize}

\section{Telecommunications Tasks Description}
The objective of this study is to evaluate the effectiveness of LLMs for NLP tasks that are specific to the telecommunications domain. We focus on $3$ distinct and crucial NLP tasks in this field, namely, text classification, summarization, and question answering, using $3$ benchmark datasets. We assess the performance of four models: Llama-$2$-$7$B \cite{touvron2023llama}, Falcon-$7$B \cite{almazrouei2023falcon}, Mistral-$7$B \cite{jiang2023mistral}, and Zephyr-$7$B-$\beta$ \cite{tunstall2023zephyr}. Given that it is not required to fine-tune the base architecture for inference, our emphasis is on creating a zero-shot learning environment for these models. We also provide the state-of-the-art results for each task as a benchmark for comparison. In this section, we introduce the benchmark telecommunications text-processing tasks that are the subject of our study. We further elaborate on these tasks and the corresponding benchmark datasets used for each task. A summary of the datasets for each of the tasks is shown in \tableautorefname~\ref{tab:dataset}.

\begin{table*}
\centering
\scriptsize
\begin{tabular}{|p{1.5cm}|p{1cm}|p{6.5cm}|p{7cm}|}
\hline
\textbf{Dataset} & \textbf{Sample} & \textbf{Instance Example} & \textbf{Instruction Prompt} \\
\hline
SPEC5G-Classification \cite{karim2023spec5g} & 2041 &\textbf{[Non-security]} An Elementary Procedure is a unit of interaction between the CN (CBC) and the RNC. \textbf{[Security]} The set of allowed range classes for each Application ID is stored in the ProSe Function Application registration procedure completion by the UE. \textbf{[Undefined]} This field indicates the Location Area Code of the reference BTS. & You are an assistant skilled in classifying text into any of the three categories - Non-Security, Security, or Undefined. Your task is to classify the given text into any of these categories. Given a text, use your capabilities to classify it into any of these three network security categories - Non-Security, Security, or Undefined. The security-text classification task involves automatically identifying texts within the 5G specifications that specify important security properties. These properties are essential for formal verification and other testing methodologies. Your response should only contain the name of the category. You must not response any extra text before or after it. Text: \textbf{[Sample]} \\
\hline
SPEC5G-Summarization \cite{karim2023spec5g} & 713 & \textbf{[Paragraph]} An active UL bandwidth part (BWP) is indicated by higher layers for a PUSCH transmission scheduled by a RAR UL grant. For determining the frequency domain resource allocation for the PUSCH transmission within the active UL BWP UE does follow below mentioned steps. In both cases, despite the start RB may be different, the maximum no. of RBs for frequency domain resource allocation always equals the number of RBs in the initial UL BWP. The frequency domain resource allocation is by uplink resource allocation type 1. \textbf{[Summary]} For determining the frequency domain resource allocation for the PUSCH transmission within the active UL, UE processes the frequency domain resource assignment fields. & You are an assistant skilled in summarizing text. Given a paragraph, use your capabilities to summarize it. Focus on extracting and highlighting the key points, main ideas, and essential details. Your summary should be concise, clear, and informative, maintaining the core message of the paragraph while removing any redundant or non-essential information. Aim to present the summarized content in a way that is easily understandable and retains the original intent and context of the paragraph. The summary should be 2-3 lines. Paragraph: \textbf{[Sample]} \\
\hline
TeleQnA \cite{maatouk2023teleqna} & 10000 & \textbf{[Question]} What are the main criteria used to validate that a sequence of numbers is random? 1. Independence and randomness 2. Uniformity and randomness 3. Independence and uniformity 4. Uniformity and unpredictability 5. Independence and unpredictability \textbf{[Answer]} 3 & You are an assistant skilled in selecting the most appropriate answer from a set of given options for each question. Your task is to read the question and the options, then choose the best answer. Given a question, select the most appropriate answer from the provided options. The response should be the number of the option you believe is correct, such as 1, 2, 3, 4 or 5. The response should consist solely of the option number (an integer) and must not include any extra text before or after it. Question: \textbf{[Sample]} \\
\hline
\end{tabular}
\caption{Dataset descriptions with instruction prompts for classification, summarization, and question answering tasks.}
\label{tab:dataset}
\end{table*}

\subsection{Text Classification}
Text classification involves assigning a specific category to a given input text. We conducted experiments with the SPEC5G—Classification dataset \cite{karim2023spec5g} to assess the text classification abilities of LLMs within the telecommunications domain. The dataset was primarily sourced from the 3GPP website, a consortium that produces reports and specifications that define cellular telecommunications technologies. 3GPP offers a plethora of meeting minutes, technical reports, and technical specifications. A significant part of the dataset was obtained from the 3GPP FTP server. However, the standard documentation often includes noisy data, which poses challenges for automated NLP tasks. To mitigate this, the authors carried out extensive preprocessing and data scraping from blogs and internet forums.
The authors then randomly selected and annotated $2401$ sentences from the SPEC5G dataset for multi-class classification. The data was divided into three classes: non-security, security, and undefined, and labeled by $9$ domain experts. The final classification dataset is somewhat imbalanced, with the class with the most samples (Non-Security: $1326$) being approximately three times larger than the class with the fewest samples (Undefined: $461$).
\subsection{Text Summarization}
Text summarization is the process of distilling the original text into a more comprehensible version while preserving the essence of the initial content.
This technique proves beneficial for non-native speakers
and readers without specialized knowledge.
In the context of 5G standard documents, summarization serves as a valuable tool for developers and practitioners, enabling them to grasp the overarching concept of the specification—a task that could otherwise be significantly time-intensive.
Experiments were carried out utilizing the SPEC5G—Summarization dataset \cite{karim2023spec5g} to evaluate the proficiency of LLMs in performing text summarization tasks specific to the telecommunications sector.
To curate the dataset, the authors selected $1500$ random articles from the SPEC5G dataset, which comprises of $17GB$ of text data from specification releases and web portals. An article is characterized as a sequential aggregation of sentences. To ensure semantic coherence among the sentences in each article, an additional round of manual processing was applied. The final refined dataset comprised $713$ articles, each containing $1-12$ sentences. Subsequently, the dataset was annotated by domain experts, with each annotation being a condensed set of sentences that encapsulates the essence of the article.

\subsection{Question Answering}
This task involves Multiple-Choice Questions (MCQs), where each question is accompanied by a variable number of options (up to $5$). The aim is to determine whether the correct answer can be deduced from these options. It is important to note that for each question, only one of these options is correct. Additionally, the LLMs are also evaluated in a traditional question-answering task manner, where the options are not provided. The LLM is instructed to provide the correct response based solely on the question. For this task, we assess the performance of LLMs on the TeleQnA dataset \cite{maatouk2023teleqna}. The authors curated a dataset consisting of $10,000$ question-answer pairs from diverse telecommunications-related sources, including technical standards documents, research materials, and telecom lexicons, and preprocessed for compatibility with LLMs. Two GPT-3.5 \cite{NEURIPS2020_1457c0d6} LLM agents, a generator, and a validator, were used to create and assess MCQs. Post-processing involves filtering, shuffling answer options, and mapping acronyms. Telecom experts reviewed the questions, options, and explanations for accuracy and relevance. The dataset was further refined through a clustering process and a secondary human validation step to eliminate redundancy.
\section{Methodology}
\subsection{Prompt Design}
For any given test sample, denoted as $X$, we create a task instruction, $T$, and combine it with the test sample to form the prompt, $P$. This prompt, $P$, is then used as input to generate a response, $R$. We illustrate how we construct the prompt, $P$, for various tasks. Sample prompts for each of the tasks are delineated in \tableautorefname~\ref{tab:dataset}.
For text classification, the goal is to categorize the provided sample text. The task instruction contains all potential class names, and the LLM is instructed to choose from the provided class labels. To clarify, the task at hand is not multi-label classification, but rather multi-class classification, implying that a single sample can only be associated with one of the possible classes. Furthermore, the LLM is instructed to output solely class labels, excluding any other text before or after it, which aids in automating the evaluation process. For the summarization task, the sample includes a paragraph that needs to be summarized. The LLM is instructed to generate a summary of the given paragraph in $2$ to $3$ sentences, encapsulating the key points, main ideas, and crucial details. The response is then assessed using Rouge scores. For MCQ-type question-answering, the sample includes both the question and the options. The LLM is instructed to choose any of the options as the correct answer. Suppose the number of samples for question $Q$ is $N$. In this case, the options in the sample are numbered from $1$ to $N$. Therefore, the LLM is instructed to provide only the option number that it deems correct for the given question, $Q$. The outputs are then evaluated in an automated manner. For the generic question-answering task, the sample only contains the question without the options. The LLMs are instructed to provide a single-sentence response and the response is evaluated against the correct option using Rouge and Bleu scores.
\subsection{Models}
\subsubsection{Llama-$2$-$7$B}
Llama-$2$, an upgrade to Llama, introduces substantial advancements in architecture and pretraining, including a more rigorous data cleaning, a $40\%$ expanded pretraining corpus, a context length that is twice as long, and Grouped-Query Attention (GQA) mechanism.
GQA facilitates efficient attention computation to handle the increased size and complexity of the Llama-$2$ models. The training of the Llama-$2$ models involves supervised fine-tuning and Reinforcement Learning with Human Feedback (RLHF).
RLHF aligns the model with human preferences, using binary comparisons and preference data that focus on helpfulness and safety. Context distillation further refines the RLHF by generating safer responses using safety preprompts, thereby improving performance and safety.
Moreover, Llama-$2$ models excel in performance and safety, surpassing open-source models and equaling some closed-source ones. They ensure secure AI deployment through red teaming, iterative evaluations, and safety-specific data annotation.

\subsubsection{Falcon-$7$B}
The Falcon series LLMs, distinguished by their superior performance and scalability, are trained on an extensive web dataset known as RefinedWeb.
It incorporates advanced features such as multi-query attention
and rotary positional embeddings
to augment efficiency. The multi-query attention mechanism of the transformer neural sequence model diminishes memory bandwidth requirements during incremental decoding.
Falcon series LLMs use FlashAttention, a fast, memory-efficient attention algorithm for Transformers. It minimizes memory exchanges between the GPU's high bandwidth memory and on-chip SRAM using tiling, speeding up training, enabling longer contexts, and improving model quality and task performance.

\subsubsection{Mistral-$7$B}
This model utilizes advanced architectural features to enhance its performance and efficiency. It includes GQA
and sliding window attention
for faster inference and effective sequence management, and techniques like rolling buffer cache, pre-fill and Chunking for better memory management. The innovative and efficient Mistral-$7$B outperforms models like LLaMA-$2$-$13$B and LLaMA-$1$-$34$B in tasks like code, math, and reasoning, despite its size. However, its knowledge benchmark scores are slightly lower due to fewer parameters.

\subsubsection{Zephyr-$7$B-$\beta$}
This model is a refined version of Mistral-$7$B. It was trained on a combination of synthetic and publicly available datasets using direct preference optimization. The initial fine-tuning was performed on a filtered and preprocessed version of the UltraChat dataset,
which includes a wide variety of synthetic dialogues produced by ChatGPT. The authors claimed that eliminating the inherent alignment of these datasets enhanced the model's performance on MT Bench.
Specifically, Zephyr-$7$B-$\beta$ outperforms larger open models like Llama-$2$-Chat-$70$B in several MT-Bench categories. However, for more intricate tasks such as coding and mathematics, this model falls short of proprietary models.

\section{Results and Discussion}
Based on our prompt-based zero-shot learning experiments, we report the performance of LLMs for different tasks and compare their performance with
the current state-of-the-art results. 
\subsection{Text Classification}
\begin{figure}
    \centering
    \includegraphics[width=0.9\linewidth]{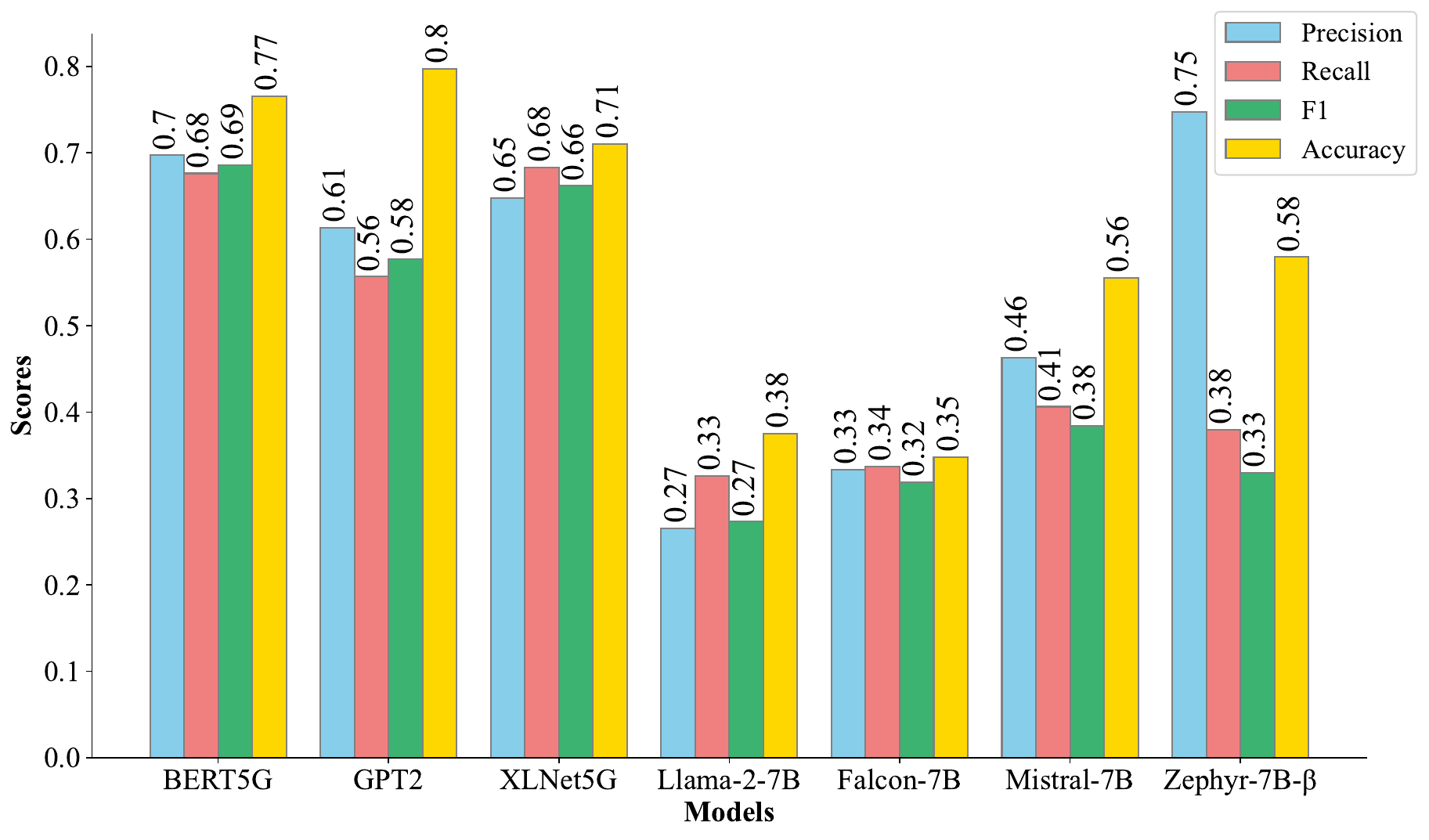}
    \caption{Text classification results for different models.}
    \label{fig:classification}
\end{figure}

In our study on the SPEC5G-Classification dataset, it was observed that the Mistral and Zephyr LLMs outperformed their counterparts (refer to \tableautorefname~\ref{tab:classification_classwise}). The performance of these two models was strikingly similar, which can be attributed to the fact that Zephyr is a fine-tuned version of the Mistral model. Conversely, the Llama-2 and Falcon models demonstrated subpar performance in the classification task. Among the four LLMs evaluated, Zephyr emerged as the top performer with an accuracy of $57.93\%$, followed closely by Mistral. Falcon, however, recorded the lowest accuracy at $34.78\%$. Upon manual inspection of the responses generated by these LLMs, it was found that the decline in performance of Falcon and Llama-$2$ was primarily due to their inability to adhere to prompt instructions. This led to incorrectly formatted responses and subsequent errors during automated evaluation. A comparative analysis with state-of-the-art supervised models (refer to \figureautorefname~\ref{fig:classification}) revealed that the zero-shot performance of LLMs falls short of the performance exhibited by supervised models in text classification tasks. The performance of LLMs falls short of supervised models, primarily because supervised models require significant training data to fine-tune on a downstream task. These downstream tasks are discriminative rather than generative, which eliminates the possibility of incoherent responses. It should be noted that the BERT5G and XLNet5G models are pre-trained on the SPEC5G \cite{karim2023spec5g} dataset.

\begin{table}
\centering
\scriptsize
\begin{tabular}{|c|c|p{0.95cm}|p{0.7cm}|p{0.65cm}|p{0.85cm}|}
\hline
\textbf{Model} & \textbf{Class} & \textbf{Precision} & \textbf{Recall} & \textbf{F1} & \textbf{Samples} \\
\hline
\multirow{3}{*}{Llama-2-7B} & Non-Security & \cellcolor{PineGreenDarkest}0.5502 & \cellcolor{PineGreenMedium}0.4216 & \cellcolor{PineGreenDarkest}0.4774 & \cellcolor{PineGreenDarkest}1326 \\
 & Security & \cellcolor{PineGreenMedium}0.2469 & \cellcolor{PineGreenDarkest}0.5570 & \cellcolor{PineGreenMedium}0.3422 & \cellcolor{PineGreenMedium}614 \\
 & Undefined & \cellcolor{PineGreenLight}0.0000 & \cellcolor{PineGreenLight}0.0000 & \cellcolor{PineGreenLight}0.0000 & \cellcolor{PineGreenLight}461 \\
\hline
\multirow{3}{*}{Falcon-7B} & Non-Security & \cellcolor{PineGreenDarkest}0.5302 & \cellcolor{PineGreenMedium}0.3439 & \cellcolor{PineGreenDarkest}0.4172 & \cellcolor{PineGreenDarkest}1326 \\
 & Security & \cellcolor{PineGreenMedium}0.2605 & \cellcolor{PineGreenDarkest}0.4658 & \cellcolor{PineGreenMedium}0.3341 & \cellcolor{PineGreenMedium}614 \\
 & Undefined & \cellcolor{PineGreenLight}0.2099 & \cellcolor{PineGreenLight}0.2017 & \cellcolor{PineGreenLight}0.2058 & \cellcolor{PineGreenLight}461 \\
\hline
\multirow{3}{*}{Mistral-7B} & Non-Security & \cellcolor{PineGreenDarkest}0.5862 & \cellcolor{PineGreenDarkest}0.8311 & \cellcolor{PineGreenDarkest}0.6875 & \cellcolor{PineGreenDarkest}1326 \\
 & Security & \cellcolor{PineGreenMedium}0.4563 & \cellcolor{PineGreenMedium}0.3485 & \cellcolor{PineGreenMedium}0.3952 & \cellcolor{PineGreenMedium}614 \\
 & Undefined & \cellcolor{PineGreenLight}0.3462 & \cellcolor{PineGreenLight}0.0390 & \cellcolor{PineGreenLight}0.0702 & \cellcolor{PineGreenLight}461 \\
\hline
\multirow{3}{*}{Zephyr-7B-$\beta$} & Non-Security & \cellcolor{PineGreenLight}0.5731 & \cellcolor{PineGreenDarkest}0.9721 & \cellcolor{PineGreenDarkest}0.7211 & \cellcolor{PineGreenDarkest}1326 \\
 & Security & \cellcolor{PineGreenMedium}0.6689 & \cellcolor{PineGreenMedium}0.1645 & \cellcolor{PineGreenMedium}0.2641 & \cellcolor{PineGreenMedium}614 \\
 & Undefined & \cellcolor{PineGreenDarkest}1.0000 & \cellcolor{PineGreenLight}0.0022 & \cellcolor{PineGreenLight}0.0043 & \cellcolor{PineGreenLight}461 \\
\hline
\end{tabular}
\caption{Class-wise classification results for different models. \textit{The better performing classes in each model are indicated by a deeper cell color.}}
\label{tab:classification_classwise}
\end{table}

\subsection{Text Summarization}
\begin{figure}
    \centering
    \includegraphics[width=0.9\linewidth]{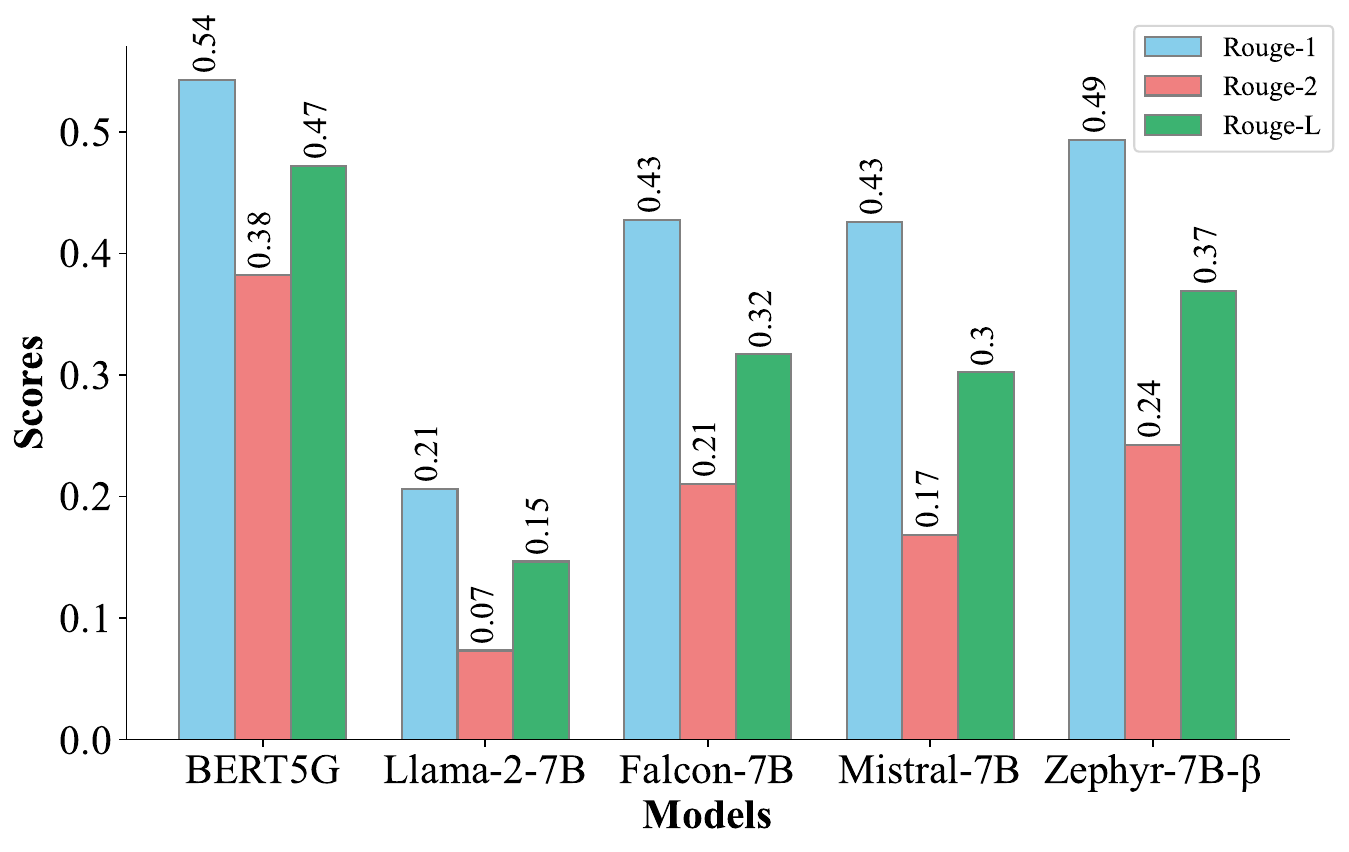}
    \caption{Summarization results for different models.}
    \label{fig:summarization}
\end{figure}

We provide our results on the SPEC5G-Summarization dataset and compare with the work of Karim et al. \cite{karim2023spec5g} in \figureautorefname~\ref{fig:summarization}. The evaluation metrics employed are Rouge-1, Rouge-2, and Rouge-L scores. The results indicate that BERT5G, pre-trained on SPEC5G \cite{karim2023spec5g}, outperforms other LLMs across all three metrics with scores of 0.54 (Rouge-1), 0.38 (Rouge-2), and 0.47 (Rouge-L). In our experimentation with various LLMs, Zephyr demonstrated superior performance across all evaluation metrics. However, the zero-shot performance is still subpar compared to a task-specific pre-trained language model, BERT5G. Zephyr’s superior Rouge-1 score, relative to other LLMs, signifies its enhanced ability to identify crucial unigrams within the summary. Furthermore, Zephyr’s elevated Rouge-2 score suggests its proficiency in recognizing significant bigrams that contribute to the summary’s meaning. The commendable Rouge-L score of Zephyr implies its capacity to discern extended word sequences present in the reference summary, thereby demonstrating its potential in understanding long-term dependencies. The performances of Falcon and Mistral are observed to be on par with those of Zephyr. 

\subsection{Question Answering}
\begin{figure}
    \centering
    \includegraphics[width=0.9\linewidth]{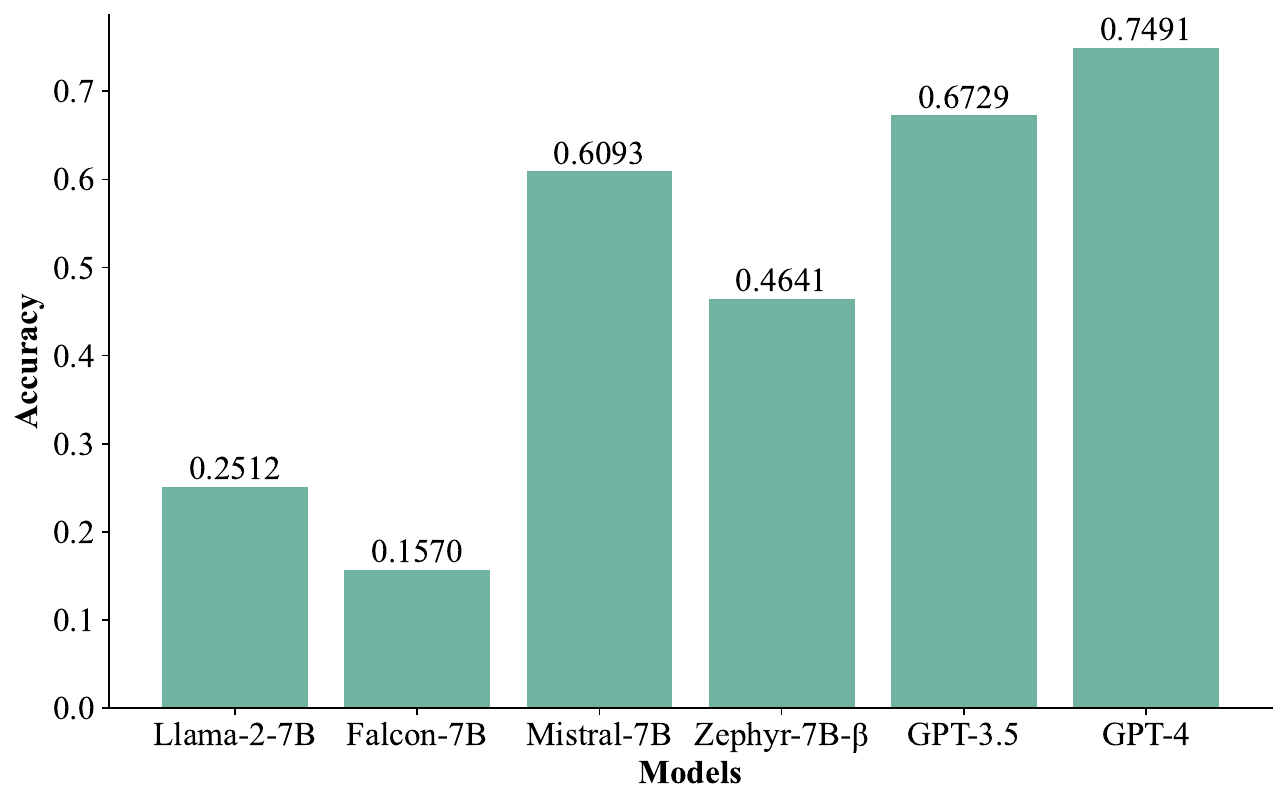}
    \caption{MCQ-type question-answering accuracy scores.}
    \label{fig:question}
\end{figure}

In our study, we assessed the performance of LLMs on the TeleQNA dataset, considering both traditional and MCQ settings. Our results were compared with the findings of Maatouk et al. \cite{maatouk2023teleqna} for MCQ-type question-answering tasks.
Mistral demonstrated an accuracy of 60.93\% in MCQ-type question-answering, a figure that is comparable to the $67.11\%$ and $74.91\%$ accuracy achieved by GPT-3.5 and GPT-4 respectively \cite{maatouk2023teleqna} (refer to \figureautorefname~\ref{fig:question}). Given that GPT-3.5 has nearly $175$ billion and GPT-4 has nearly $1.76$ trillion parameters, the performance of Mistral, with only $7$ billion parameters, is particularly impressive. Zephyr’s performance closely follows that of Mistral. However, the performances of Llama-$2$ and Falcon were found to be subpar. Considering that each question typically had 4-5 options, the performance of these two LLMs essentially equates to random guessing. The class-wise accuracy of these LLMs is shown in \tableautorefname~\ref{tab:classwise_qa}. We conducted additional assessments of these LLMs in a conventional question-answering context using Rouge and Bleu metrics (refer to \tableautorefname~\ref{tab:combined_results}). Nevertheless, when the prompt only contains the question without any options, the LLMs' performance is below average, and they all have difficulty producing responses that closely match the ground truths. Their performances are similar although the performance of Llama-$2$ is slightly better than others.

\begin{table}
\centering
\scriptsize
\begin{tabularx}{\columnwidth}{|l|X|X|X|X|}
\hline
\textbf{Category} & \textbf{Llama-$2$-$7$B} & \textbf{Falcon-$7$B} & \textbf{Mistral-$7$B} & \textbf{Zephyr-$7$B-$\beta$} \\ \hline
{Lexicon} & \cellcolor{PineGreenLighter}0.218 & \cellcolor{PineGreenDarkest}0.168 & \cellcolor{PineGreenDarkest}0.692 & \cellcolor{PineGreenDarkest}0.536 \\
{Research Overview} & \cellcolor{PineGreenLight}0.245 & \cellcolor{PineGreenLight}0.157 & \cellcolor{PineGreenDark}0.659 & \cellcolor{PineGreenMedium}0.478 \\ 
{Research Publications} & \cellcolor{PineGreenMedium}0.247 & \cellcolor{PineGreenDark}0.160 & \cellcolor{PineGreenMedium}0.633 & \cellcolor{PineGreenDark}0.485 \\ 
{Standards Overview} & \cellcolor{PineGreenDark}0.263 & \cellcolor{PineGreenMedium}0.158 & \cellcolor{PineGreenLight}0.588 & \cellcolor{PineGreenLight}0.449 \\ 
{Standards Specifications} & \cellcolor{PineGreenDarkest}0.270 & \cellcolor{PineGreenLighter}0.147 & \cellcolor{PineGreenLighter}0.497 & \cellcolor{PineGreenLighter}0.393 \\ \hline
\end{tabularx}
\caption{Accuracy for Different Question-Answering Categories. \textit{Within each column, cells with higher values are emphasized with progressively deeper shades of color.}}
\label{tab:classwise_qa}
\end{table}

A summary of our experimental results for different tasks is shown in \tableautorefname~\ref{tab:combined_results}.

\begin{table*}
\centering
\scriptsize
\begin{tabular}{|c|c|c|c|c|c|c|c|c|c|c|c|c|}
\hline
\textbf{Model} & \multicolumn{3}{c|}{\textbf{Summarization}} & \multicolumn{5}{c|}{\textbf{Question Answering}} & \multicolumn{4}{c|}{\textbf{Classification}} \\
\cline{2-13}
& \textbf{R-1} & \textbf{R-2} & \textbf{R-L} & \textbf{Acc.} & \textbf{R-1} & \textbf{R-2} & \textbf{R-L} & \textbf{Bleu} & \textbf{Precision} & \textbf{Recall} & \textbf{F1} & \textbf{Acc.} \\
\hline
Llama-$2$-$7$B & 
\cellcolor{PineGreenLight}0.2064 & \cellcolor{PineGreenLight}0.0733 & \cellcolor{PineGreenLight}0.1465 & \cellcolor{PineGreenMedium}0.2512 & \cellcolor{PineGreenDarkest}0.1301& \cellcolor{PineGreenDark}0.0285 & \cellcolor{PineGreenDarkest}0.1173 & \cellcolor{PineGreenDark}0.0179 & \cellcolor{PineGreenLight}0.2657 & \cellcolor{PineGreenLight}0.3262 & \cellcolor{PineGreenLight}0.2732 & \cellcolor{PineGreenMedium}0.3753 \\
Falcon-$7$B & 
\cellcolor{PineGreenDark}0.4278 & \cellcolor{PineGreenDark}0.2102 & \cellcolor{PineGreenDark}0.3169 & \cellcolor{PineGreenLight}0.1570 & \cellcolor{PineGreenLight}0.1131 & \cellcolor{PineGreenLight}0.0233 & \cellcolor{PineGreenLight}0.1028 & \cellcolor{PineGreenMedium}0.0154 & \cellcolor{PineGreenMedium}0.3335 & \cellcolor{PineGreenMedium}0.3371 & \cellcolor{PineGreenMedium}0.3190 & \cellcolor{PineGreenLight}0.3478 \\
Mistral-$7$B & 
\cellcolor{PineGreenMedium}0.4259 & \cellcolor{PineGreenMedium}0.1684 & \cellcolor{PineGreenMedium}0.3022 & \cellcolor{PineGreenDarkest}0.6093 & \cellcolor{PineGreenDark}0.1265 & \cellcolor{PineGreenDarkest}0.0303 & \cellcolor{PineGreenDark}0.1145 & \cellcolor{PineGreenDarkest}0.0188 & \cellcolor{PineGreenDark}0.4629 & \cellcolor{PineGreenDarkest}0.4062 & \cellcolor{PineGreenDarkest}0.3843 & \cellcolor{PineGreenDark}0.5556 \\
Zephyr-$7$B-$\beta$ & 
\cellcolor{PineGreenDarkest}0.4933 & \cellcolor{PineGreenDarkest}0.2425 & \cellcolor{PineGreenDarkest}0.3693 & \cellcolor{PineGreenDark}0.4641 & \cellcolor{PineGreenMedium}0.1199 & \cellcolor{PineGreenMedium}0.0256 & 
\cellcolor{PineGreenMedium}0.1054 & \cellcolor{PineGreenLight}0.0130 & \cellcolor{PineGreenDarkest}0.7473 & \cellcolor{PineGreenDark}0.3796 & \cellcolor{PineGreenDark}0.3298 & \cellcolor{PineGreenDarkest}0.5793 \\
\hline
\end{tabular}
\caption{Performance metrics in summarization, question answering, and classification across various models. \textit{Within each column, cells with higher values are emphasized with progressively deeper shades of color.}}
\label{tab:combined_results}
\end{table*}

\section{Error Analysis}
We observed interesting cases where LLMs produced incoherent responses to instructional prompts. This section details our findings for each task.

\subsection{Text Classification}
The LLMs were instructed to produce only the class label from a sample. Given that the class label is a single term, we limited the generation of new tokens to three. The same prompt was given to all LLMs. Llama-$2$ and Zephyr adhered to the instructions successfully, but the other two LLMs did not. For each dataset instance, the responses from Llama-$2$ and Zephyr consisted solely of the class label, with no preceding or following tokens. On the other hand, Falcon and Mistral's responses were inconsistent in some instances. Rather than generating the class label, they produced various strings like `Understood. I will', `Based on the', etc. Falcon also generated some telecommunications-related terms. An interesting observation was that despite having three classes - `Non-security', `Security', and `Undefined', Llama-$2$ and Zephyr never classified any sample as `Undefined'. We hypothesize that incorporating the class definitions could enhance the classification performance of all models.

\subsection{Text Summarization}
In the summarization task, the performance of Llama-$2$ was notably inferior to the other three LLMs. Upon examining the responses, we found that Llama-$2$'s responses were excessively lengthy for a summarization task, while the other LLMs typically limited their summaries to 2 to 5 sentences. Furthermore, Llama-$2$ produced repetitive text such as `I understand your request, ...', `Hello! As an ethical and knowledgeable assistant ...', etc. at the start of each response. Additionally, for Llama-$2$, the summarized versions were nearly the same length as the original samples.

\subsection{Question Answering}
We evaluated the LLMs in two different settings for this task: MCQ-type and traditional. For MCQ-type question-answering, we limited the generation of new tokens to one since the LLMs were instructed to generate the best option number only. Since this instruction resembles the classification task, the observations are quite similar. Mistral and Llama-$2$ generated coherent responses to the instruction prompts, whereas the other two LLMs occasionally generated random tokens instead of the correct option number. For instance, we observed that the randomly generated tokens include `The', `Which', `Your', etc. In a traditional question-answering setting where the options are not provided in the prompt, we observed that the length of answers for the LLMs is generally longer than the ground truth. Llama-$2$ generated the longest responses among the LLMs. However, although some answers are theoretically correct, due to the longer responses, all LLMs were penalized in the Rouge and Bleu scores. For this reason, a human evaluation is necessary for traditional question-answering tasks.

\section{Limitations}
A significant part of this research is to evaluate the effectiveness of LLMs in NLP tasks within the telecommunications sector. We selected $4$ specific LLMs for this purpose, each with roughly $7$ billion parameters. We intentionally excluded larger LLMs with significantly more parameters, such as GPT-$3.5$, GPT-$4$, and so on. Our goal was to evaluate performance under resource-constrained conditions where the models need to operate locally. We hypothesized that in most scenarios, it is impractical to run extremely large models locally without assistance or necessity from other entities. Therefore, the results we have presented are from models whose weights are publicly accessible and can be deployed locally with a single, decent GPU. Whether the lack of benchmarks from large-scale LLMs is considered a limitation of this study is subjective and depends on the individual reader's perspective. Another important aspect is that we did not employ any prompt optimization methods or evaluation for $k$-shot inference where $k>0$. The prompts were crafted with the sole intention that the LLM should respond in a particular format to enable automated assessment, and the prompts should not be specific to any domain. Despite the demonstrated effectiveness of prompt optimization and $k$-shot in NLP tasks, we aimed for our findings to be as universal as possible to allow a fair comparison of the LLMs' performances.

\section{Conclusion and Future Work}
This study evaluates the strengths and weaknesses of LLMs in $3$ benchmark tasks in telecommunications. To the best of our knowledge, this is the first comprehensive assessment of LLMs in this domain. Our empirical results indicate that Mistral and Zephyr demonstrated remarkable zero-shot performance across all tasks, while Llama-$2$ and Falcon's performances were subpar in most experiments. Despite this, the results reveal that the performance of zero-shot LLMs is yet to match that of state-of-the-art supervised fine-tuned models. We also conclude that while smaller models like Zephyr or Mistral perform admirably, larger models such as GPT-$3.5$ or GPT-$4$ significantly surpass these smaller models. Future research should explore the performance of $k$-shot LLMs, prompt optimization, and efficient fine-tuning techniques.
Additionally, it is crucial to examine potential biases, ethical issues, and the risk of misinformation generation associated with LLMs. For tasks like summarization or question-answering, human evaluation proves to be more effective than automated evaluation using Rouge or Bleu scores. Hence, future investigations should integrate human evaluation into their framework.
\bibliographystyle{IEEEtran}
\bibliography{ref}

\end{document}